\pdfoutput=1

\documentclass[11pt]{article}

\usepackage[final]{acl}

\usepackage{times}
\usepackage{latexsym}
\usepackage{pdfpages}

\usepackage{amsmath}
\usepackage{amssymb}

\usepackage{hyperref}

\usepackage[small,compact]{titlesec}
\usepackage[normalem]{ulem}
\useunder{\uline}{\ul}{}
\usepackage{booktabs}       
\usepackage{enumitem}

\usepackage{float}

\usepackage[T1]{fontenc}

\usepackage[utf8]{inputenc}

\usepackage{microtype}

\usepackage{makecell}
\newcolumntype{P}[1]{>{\centering\arraybackslash}p{#1}}

\title{Improving Entity Disambiguation by Reasoning over a Knowledge Base}

\author{\centering\begin{tabular}{P{50mm}P{50mm}P{50mm}}
        {\textbf{Tom Ayoola$^*$}} & {\textbf{Joseph Fisher$^*$}} & \textbf{Andrea Pierleoni} \\
       \multicolumn{3}{c}{\textnormal{Amazon Alexa AI}}                                     \\
       \multicolumn{3}{c}{\textnormal{Cambridge, UK}}                                        \\
       \multicolumn{3}{c}{\textnormal{\texttt{\{tayoola, fshjos, apierleo\}@amazon.com}}}   
    \end{tabular}}

\begin{document}
\maketitle
\renewcommand*{\thefootnote}{\fnsymbol{footnote}}
\footnotetext[1]{Tom and Joseph contributed equally to this work.}
\renewcommand*{\thefootnote}{\arabic{footnote}}
\begin{abstract}
Recent work in entity disambiguation (ED) has typically neglected structured knowledge base (KB) facts, and instead relied on a limited subset of KB information, such as entity descriptions or types.
This limits the range of contexts in which entities can be disambiguated.
To allow the use of all KB facts, as well as descriptions and types, we introduce an ED model which links entities by reasoning over a symbolic knowledge base in a fully differentiable fashion. Our model surpasses state-of-the-art baselines on six well-established ED datasets by 1.3 F1 on average. By allowing access to all KB information, our model is less reliant on popularity-based entity priors, and improves performance on the challenging ShadowLink dataset (which emphasises infrequent and ambiguous entities) by 12.7 F1.


\end{abstract}

\section{Introduction}

Entity disambiguation (ED) is the task of linking mentions of entities in text documents to their corresponding entities in a knowledge base (KB). Recent ED models typically use a small subset of KB information (such as entity types or descriptions) to perform linking. These models have strong performance on standard ED datasets, which consist mostly of entities that appear frequently in the training data.

However, ED performance deteriorates for less common entities, to the extent that many recent models are outperformed by outdated feature engineering-based ED systems on datasets that focus on challenging or rare entities \cite{provatorova-etal-2021-robustness}. This suggests models over-rely on prior probabilities, which are either implicitly learned or provided as features, rather than make effective use of the mention context. One reason for this is that the subset of KB information used by the models is not enough to discriminate between similar entities in all contexts, meaning the model has to fall back on predicting the most popular entity. Another explanation for the performance drop is that less common entities are prone to missing or inconsistent KB information (e.g. they may not have a description), which is problematic for models which rely on a single source of information. To illustrate, we find that 21\% of the 25\% least popular\footnote{We use the number of KB facts where the entity is the subject entity as a proxy for popularity, and only consider entities with an English Wikipedia page.} entities in Wikidata have neither an English description nor any entity type\footnote{See for example \href{ https://www.wikidata.org/wiki/Q5017238}{Q5017238}.}, leaving no mechanism for models which rely on these two sources of information alone to disambiguate them (other than their label).\footnote{Conversely, 100\% of the 25\% most popular entities in Wikidata have either a description or type.} Over half of these entities have at least one KB fact (e.g. [Cafe Gratitude], [headquarters location], [San Francisco]); so by including KB facts the percentage of the least popular entities with no information aside from a label drops from 21\% to 8\%.

\begin{figure}[h]
    \centering
    \includegraphics[page=1,width=0.45\textwidth]{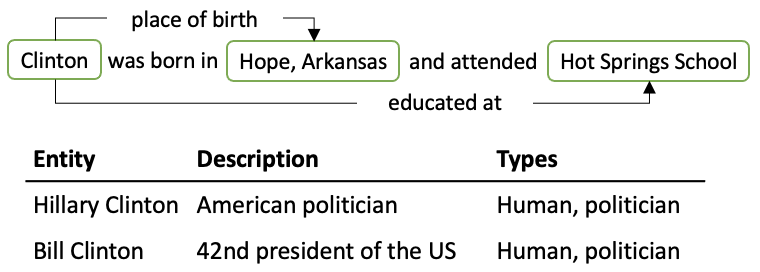}
    \caption{Example of a sentence where fine-grained KB information is required for entity disambiguation.}
    \label{examplesentence}
\end{figure}



In light of this, we introduce an ED model which has access to entity types and descriptions, and all KB facts. By using a larger variety of information, our model is more robust to missing KB information, and is able to disambiguate entities in a broader range of contexts without relying on entity priors. Figure \ref{examplesentence} shows an example sentence where there is insufficient information in the entity descriptions and types to disambiguate the mention, \emph{Clinton}. Fine-grained KB information, such as facts about the birthplace or education of candidate entities, is required.


\begin{figure*}[htp]
    \centering
    \includegraphics[page=1,width=0.8\textwidth]{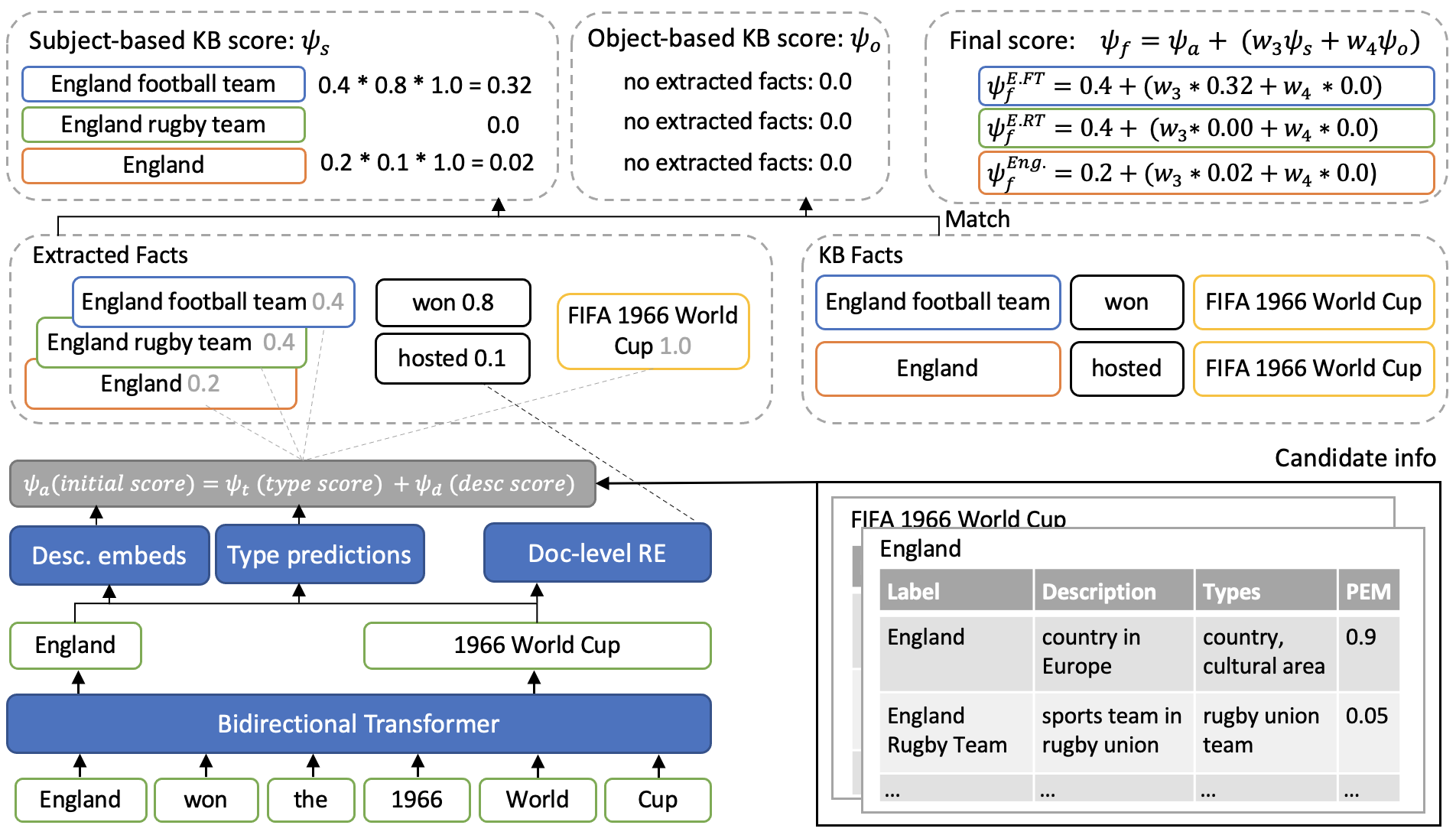}
    \caption{Our model architecture shown for a document with two mentions, \emph{England} and \emph{1966 World Cup}. The model disambiguates all entity mentions in a single pass; making use of the KB facts connecting the candidates of each mention.}
    \label{modeldiagram}
\end{figure*}

To incorporate KB facts, our model begins by re-ranking candidate entities using descriptions \cite{blink} and predicted entity types \cite{Raiman2018DeepTypeME}. We then predict, using the document context, the relations which exist between every pair of mentions in the document. For example, given the sentence in Figure \ref{examplesentence}, the model may predict that the [place of birth] relation exists between the mention \emph{Clinton} and the mention \emph{Hope, Arkansas}.\footnote{We use square brackets to denote relations and entities in the KB, and italics to represent mentions in the input text.} For this, we introduce a novel ``coarse-to-fine'' document-level relation extraction (RE) module, which increases accuracy and reduces inference time relative to the standard RE approach. Given the relation predictions, we query the KB (Wikidata in our case) for facts which exist between any of the candidate entities for the mention \emph{Clinton} and for the mention \emph{Hope, Arkansas}. In this case we would find the Wikidata fact [Bill Clinton], [place of birth], [Hope], and would correspondingly boost the scores of both the [Bill Clinton] and [Hope] entities. 
We implement this mechanism with the KB stored in a one-hot encoded sparse tensor, which makes the architecture end-to-end differentiable.


Our model surpasses state-of-the-art (SOTA) baselines on well-established ED datasets by 1.3 F1 on average, and significantly improves performance on the challenging ShadowLink dataset by 12.7 F1. In addition, the model predictions are interpretable, in that the facts used by the model to make predictions are accessible.



Our contributions are summarised as follows:
\begin{enumerate}
    \item We empirically show that using KB facts for ED increases performance above SOTA methods, which generally rely on a single source of KB information.
    \item We introduce a scalable method of incorporating symbolic information into a neural network ED model. To our knowledge, this is the first time an end-to-end differentiable symbolic KB has been used for ED.
    \item  We introduce a novel document-level relation extraction (RE) architecture which uses coarse-to-fine predictions to obtain competitive accuracy with high efficiency.  
\end{enumerate}


\section{Related Work}
Recent work on ED has primarily focused on feature-based approaches, whereby a neural network is optimised so that the representation of the correct KB entity is most similar to the mention representation, and each mention is resolved independently. The way in which the KB entities are represented varies between work. Initial work \cite{ganea-hofmann-2017-deep} learned entity embeddings directly from training examples, which performed well for entities seen during training, but could not resolve unseen entities. More recent work improved performance on common datasets by enabling linking to entities unseen during training by using a subset of KB information to represent entities, such as entity descriptions \cite{logeswaran-etal-2019-zero, wu-etal-2020-scalable} or entity types \cite{Raiman2018DeepTypeME, DBLP:conf/aaai/OnoeD20}. 


\subsection{ED with KB context}
\citet{mulang} and \citet{triplesastext} incorporate KB facts into ED models by lexicalising KB facts and appending them to the context sentence, then using a cross-encoder model to predict whether the facts are consistent with the sentence. Our model differs from this approach as we resolve entities in the document collectively rather than independently; enabling pairwise dependencies between entity predictions to be captured. Another potential limitation of the cross-encoder method is the high computational cost of encoding the long sequence length of every fact appended to the document context.
By accessing KB facts from sparse tensors, we are able to avoid this bottleneck and scale to a larger volume of facts \cite{Cohen2020Scalable}.

\subsection{ED with knowledge graph embeddings}
Graph neural networks (GNN) have been used to represent KB facts to inform ED predictions \cite{sevgili-etal-2019-improving, Ma2021-dg}. These approaches can potentially access the information in all KB facts, but are reliant on the quality of the graph embeddings, which may struggle to represent many basic semantics \cite{kgearebad} particularly for unpopular entities \cite{unpopularkge}. 


\subsection{Global ED}

There has been a series of papers which aim to optimise the global coherence of entity choices across the document \cite{hoffart-etal-2011-robust, cheng-roth-2013-relational, moro-etal-2014-entity, Pershina2015PersonalizedPR}. Our model differs from previous approaches in that the model predicts the relations which exist between mentions based on the document text and weights the coherence scores by these predictions, rather than considering coherence independently of document context. We also limit the model to pairwise coherence between mentions as opposed to global coherence for computational efficiency.

\subsection{ED with multiple modules}
The most similar work to ours is \citet{bootleg}, which achieves strong results on tail entities by introducing an ED model which uses entity embeddings, relation embeddings, type embeddings, and a KB module to link entities. A key difference to our model is the way in which KB facts are used for disambiguation. In their work, KB facts are encoded independently of the document context in which the candidate entities co-occur, whereas our model is able to leverage the relevant KB facts for the document context.


\section{Proposed Method}
\subsection{Task formulation}
Given a document $X$ with mentions, $M = \{m_1, m_2, ... m_{|M|}\}$, a KB with a set of facts $G = \{(s, r ,o) \subset E \times R \times E\}$ which express relations $r \in R$ between subject $s \in E$ and object entities $o \in E$, and a description $d_k$ for each KB entity $e_k$, the goal of ED is to assign each mention $m \in M$ the correct corresponding KB entity $e \in E$.

\subsection{Overview}
Figure \ref{modeldiagram} shows a high-level overview of our model. We use a transformer model to encode all mentions in the document in a single-pass. We use these mention embeddings both to generate initial candidate entity scores for each mention using the entity types and descriptions of KB entities and to predict relations between every pair of mentions in the document. We retrieve KB facts for every pair of mentions in the document, for each combination of candidate entities. We weight the retrieved KB facts by multiplying the initial candidate entity score for the subject entity, the predicted score for the relation, and the initial candidate entity score for the object entity. Then we generate KB scores by summing the weighted facts for each candidate entity. The final score used for ranking entities is a weighted sum of the initial score and KB score.

\subsection{Mention representation}
We encode the tokens in the document $X$ using a transformer-based model, giving contextual token embeddings $H = \{\mathbf{h_1}, \mathbf{h_2}, ..., \mathbf{h_N}\}$.\footnote{We use bold letters for vectors throughout our paper.} We obtain mention embeddings $\mathbf{m_i}$ for each mention $m_i$ by average pooling the contextualised token embeddings of the mention from the final transformer layer. This allows all mentions $M$ in the document $X$ to be encoded in a single forward pass.

\subsection{Initial entity score $\mathbf{\psi_a}$}

Initially, we score candidate entities using entity typing and description scores. We combine the two with a learned weighted sum $\psi_a$:
\begin{equation} \psi_a(c_{ik}) = w_1 \psi_t(c_{ik}) + w_2 \psi_d(c_{ik}) \end{equation}
where $c_{ik}$ is the mention-entity pair ($m_i$, $e_k$), $\psi_t$ is a scoring function based on candidate entity types, and $\psi_d$ is a scoring function based on candidate entity descriptions.


\subsubsection{Entity typing score $\mathbf{\psi_t}$}
We construct a fixed set of types $T = \{(r, o) \subset R \times E\}$ by taking relation-object pairs (r, o) from the KB $G$; for example (instance of, song). We predict an independent unnormalised score for each type $t \in T$ for every mention in the document by applying a linear layer $\mathit{FF}_1$ to the mention embedding $\mathbf{m_i}$. To compute entity scores $\psi_t$ using the predicted types, we calculate the dot product between the predicted types and the candidate entity's types binary vector $\mathbf{t_k}$.\footnote{We use 1 to indicate the presence of an entity type and 0 the absence of an entity type for our binary vector. We also follow this convention for the KB facts binary vector.} 
Additionally, we add a $P(e|m)$ (PEM score) which expresses the probability of an entity given the mention text only, and is obtained from hyperlink count statistics as in previous work \cite{Raiman2018DeepTypeME}:

\begin{equation} \psi_t(c_{ik}) = (\mathit{FF}_1(\mathbf{m_i}) \cdot \mathbf{t_k}) + P(e_k | m_i)\end{equation} 

\subsubsection{Entity description score $\mathbf{\psi_d}$}
We use a bi-encoder architecture similar to \cite{blink} but altered to encode all mentions in a sequence in a single forward pass, as opposed to requiring one forward pass per mention. We represent KB entities as:
\begin{itemize}
\item[] {[CLS] label [SEP] description [SEP]}
\end{itemize}
where ``label'' and ``description'' are the tokens of the entity label and entity description in the KB. We refer to this as $d_k$. To compute entity scores $\psi_d$ using entity descriptions, we use a separate transformer model $TR_1$ to encode $d_k$, taking the final layer embedding for the [CLS], and calculate the dot product between this embedding and the contextual mention embedding $\mathbf{m_i}$ projected by linear layer $\mathit{FF}_2$:


\begin{equation} \psi_d(c_{ik}) = \mathit{FF}_2(\mathbf{m_i}) \cdot \mathit{TR}_1(d_k)  \end{equation}

\subsection{Relation extraction}
\label{subsec:relation-extraction}
Our relation extraction layer outputs a relation score vector $\mathbf{ \hat r_{ij}} \in \mathbb{R}^{|R|}$ for each pair of mentions $m_i$ and $m_j$ in the document, where $R$ is the subset of relations chosen from the KB. 
To calculate $\mathbf{ \hat r_{ij}}$ we begin by passing $\mathbf{m_i}$ and $\mathbf{m_j}$ through a bilinear layer $B$ with output dimension 1, to predict the likelihood $\hat r_{ij}^{coarse}$ that a relation exists between mentions $m_i$ and $m_j$. 
    
\begin{equation} 
 \hat r_{ij}^{coarse} = \sigma(B(\mathbf{m_i}, \mathbf{m_j}))
\end{equation}

Note that $\hat r_{ij}^{coarse}$ is a scalar, denoting the likelihood that \textbf{any} relation exists between mention $m_i$ and $m_j$.

\begin{figure}[h]
    \centering
    \includegraphics[page=1,width=0.48\textwidth]{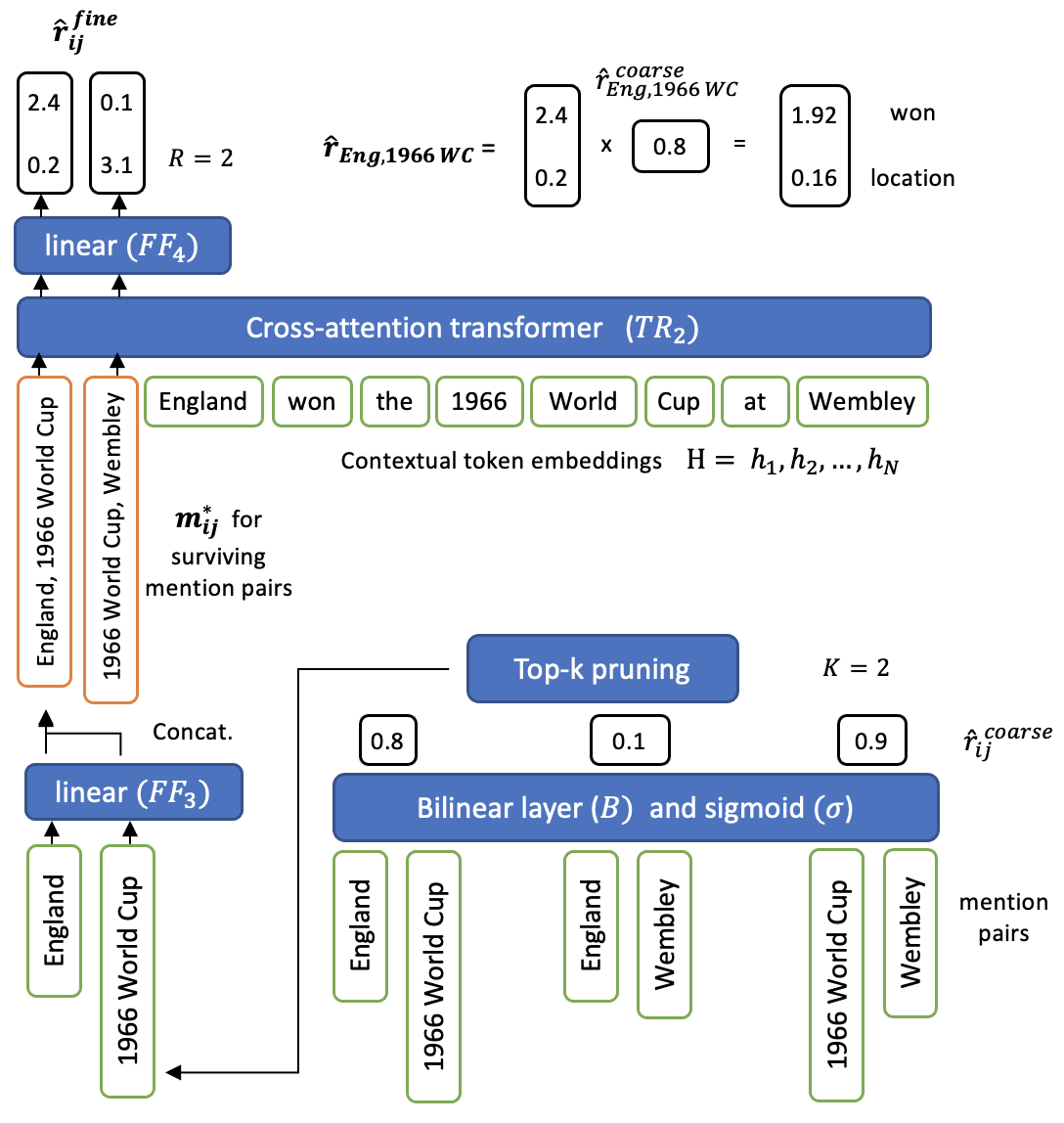}
    \caption{The model component for document-level relation extraction. $\hat r_{ij}^{coarse}$ denotes the predicted probability that any relation exists between mentions $i$ and $j$. $R$ denotes the number of relations we include in the model - set to 2 in the Figure for illustration purposes only.}
    \label{re_architecture}
\end{figure}

We then take the top-k mention pairs with the highest values of $\hat r_{ij}^{coarse}$ (in similar style to \citet{coarse2fine} who introduce a coarse-to-fine approach for coreference resolution), illustrated with $K=2$ in Figure \ref{re_architecture}. These are the pairs of mentions which the model predicts have the highest likelihood of having a relation connecting them. For the surviving mention pairs, we pass each of the two mention embeddings individually through a linear layer, $\mathit{FF}_3$, to reduce their dimension by a factor of two. This ensures that when we concatenate the two representations back together we get a representation of the mention pair $ \mathbf{m_{ij}^*}$ of the same dimension as the contextual token embeddings $H$.

\begin{equation}
    \mathbf{m_{ij}^*} = \mathit{concat}(\mathit{FF}_3(\mathbf{m_i}), \mathit{FF}_3(\mathbf{m_j}))
\end{equation}

We then pass the resulting embedding $\mathbf{m_{ij}^*}$ through a series of transformer layers $\mathit{TR}_2$, where they can attend to the contextual embeddings of the original input tokens, $H = {\{\mathbf{h_1}, \mathbf{h_2}, ..., \mathbf{h_N}\}}$. The mention-pair embeddings from the final transformer layer are passed through a linear layer $\mathit{FF}_4$ with output dimension $|R|$ to give the score that each relation exists between this mention-pair, $\mathbf{ \hat r_{ij}^{fine}}$.

\begin{equation} 
\mathbf{ \hat r_{ij}^{fine}} = \mathit{FF}_4(\mathit{TR}_2(\mathbf{m_{ij}^*}, H))) 
\end{equation}

Finally, to get $\mathbf{ \hat r_{ij}}$ we multiply the coarse layer score $\hat r_{ij}^{coarse}$ with the fine layer score $\mathbf{ \hat r_{ij}^{fine}}$, ensuring that gradients are propagated through the coarse layer during training, despite only the top-k mention pairs being passed to the fine layer. 

\begin{equation} 
\mathbf{ \hat r_{ij}} =  \hat r_{ij}^{coarse} * \mathbf{ \hat r_{ij}^{fine}}
\end{equation}

For all mention pairs outside the top-k pairs, we set $\mathbf{ \hat r_{ij}}$ to a vector of 0s. 

The relation extraction layer is trained end-to-end using the signal from the entity disambiguation loss only, and is not pretrained with any task-specific relation extraction data. To validate the effectiveness of the architecture, we include results with the RE module trained in isolation on the DOCRED RE dataset in Appendix \ref{sec:doc_level_re}.

\subsection{KB score $\mathbf{\psi_b}$}
We retrieve KB facts\footnote{Facts are efficiently retrieved by indexing into a sparse tensor.} for every mention-entity pair in the document and represent it as a 5-dimensional tensor $\mathbf{{r}}$, where $\mathbf{{r}_{ij,kn}}$ is a binary vector indicating the relations that exist in the KB between the two entities ($e_k$ and $e_n$) for mention-entity pair $c_{ik}$ and $c_{jn}$.\footnote{{The dimensions of tensor \textbf{r} are: [n\_mentions (M), n\_mentions (M), n\_candidates, n\_candidates, n\_relations (R)]}}
We weight KB facts $\mathbf{{r}}$ based on initial entity scores $\psi_a$ and relation predictions $\mathbf{\hat r}$, according their relevance to the document. To compute the KB score $\psi_b$ for a mention-entity pair, we sum KB facts where the entity (from the mention-entity pair) is the subject entity to give score $\psi_s$ and sum the KB facts where the entity is the object entity to give score $\psi_o$:

\small{
\begin{equation}
\psi_s(c_{ik}) = {\dot \psi_a(c_{ik})} \sum_{j=1}^{j \le |M|}   {\sum_{n=1}^{n \le |E|}({\mathbf{ \hat r_{ij}} \cdot \mathbf{{r}_{ij,kn}}) { {\dot \psi_a(c_{jn}) } }}}
\end{equation}
}

\small{
\begin{equation}
\psi_o(c_{ik}) = {\dot \psi_a(c_{ik})} \sum_{j=1}^{j \le |M|}   {\sum_{n=1}^{n \le |E|}({\mathbf{ \hat r_{ji}} \cdot \mathbf{{r}_{ji,nk}}) { {\dot \psi_a(c_{jn})  } }}}
\end{equation}
}

\normalsize
where $\dot \psi_a$ is the initial entity scoring function $\psi_a$ followed by the softmax function applied over the candidate entities for the given mention. We then combine the two scores with a weighted sum giving $\psi_b$:

\begin{equation}
\psi_b(c_{ik}) = w_3\psi_s(c_{ik}) + w_4\psi_o(c_{ik})
\end{equation}

Note that for computational efficiency, this scoring mechanism considers the coherence of entity predictions between pairs of mentions only, in contrast to methods which consider global coherence \citet{hoffart-etal-2011-robust}.

\subsection{Optimisation and inference}
To obtain final entity scores $\psi_f$, we add the KB scores $\psi_b$ to the initial entity scores $\psi_a$.
\begin{equation}
 \psi_f(c_{ik}) = \psi_a(c_{ik}) + \psi_b(c_{ik}) 
\end{equation}

We train our model on entity linked documents using cross-entropy loss. Our model is fully differentiable end-to-end, with the training signal propagating through all modules, including the relation extraction module. During ED inference, we take the candidate entity with the highest final entity score for each mention.

\section{Experiments}




\subsection{Standard ED}
We evaluate our model on the following well-established standard ED datasets: AIDA-CoNLL \citep{hoffart-etal-2011-robust}, MSNBC \citep{msnbc}, AQUAINT \citep{aquaint}, ACE2004 \citep{ace2004}, CWEB \citep{cweb} and WIKI \citep{wiki}. We train our model on Wikipedia hyperlinks and report \emph{InKB} micro-F1 (which only considers entities with a non-NIL entity label). To ensure fair comparisons to baselines, we use the same method to generate candidates as previous work \citep{decao2020autoregressive, le-titov-2018-improving}. Concretely, we use the top-30 entities based on entity priors (PEM) obtained by mixing hyperlink count statistics from Wikipedia hyperlinks, a large Web corpus, and YAGO.




\subsection{Long-tail and ambiguous ED}
We use the ShadowLink ED dataset \citep{provatorova-etal-2021-robustness} to evaluate our model on long-tail and ambiguous examples.\footnote{A long-tail entity is an entity that is linked to less than 56 times from other Wikipedia pages.} The dataset consists of 3 subsets. SHADOW where the correct entity is overshadowed by a more popular entity; TOP where the correct entity is the most popular entity; and TAIL where the correct entity a long-tail entity.\footnote{E.g. if the candidates and PEM scores for the mention \emph{England} were ([England (country)], 0.92) and ([England football team], 0.08) then [England (country)] would be a TOP entity, and [England football team] would be a shadow entity.} All examples in SHADOW and TOP are ambiguous, whereas TAIL has some unambiguous examples, as it is a representative sample of long-tail entities. The original dataset consists of short text snippets from Web pages, which often only include one or two mentions of entities. This limits the ability of our model to use its document-level RE module, and reason over the relationships between entities. We therefore also evaluate on the full-text version of the {SHADOW} and TOP subsets, referred to as {SHADOW-DOC} and {TOP-DOC} in the results tables.\footnote{Details in Appendix \ref{sec:dataset_details}.} The dataset consists of 1 annotated entity per document, so we run spaCy (``en\_core\_web\_lg'' model) \cite{spacy2} to identify additional mentions to allow our model and baselines to utilise other mentions to disambiguate the annotated entity mention.



\subsection{Model details}
We use Wikidata (July 2021) as our KB, restricted to entities with a corresponding English Wikipedia page. This results in 6.2M entities. We use this data to generate lookups for entity types, entity descriptions, and KB facts. We select a fixed set of 1400 relation-object pairs, based on usefulness for disambiguation, to use as our entity types (Appendix \ref{sec:model-details}). For the KB facts, we represent the top 128 relations as separate classes and collapse the remaining relations into a single class we refer to as \emph{OTHER}. Additionally, we add a special relation which exists between every entity and itself. We refer to this relation as the \emph{SAME AS} relation, and the idea behind this is to enable the model to implicitly learn coreference resolution.


\subsection{Training details}
We use Wikipedia hyperlinks (July 2021) with additional weak labels as our training dataset, which consists of approximately 100M labelled mentions. We limit candidate generation to top-30 entities based on entity priors obtained from Wikipedia hyperlink statistics.\footnote{We add weak labels by labelling spans that match the title of the page with the entity for the page.} Our model operates at the document-level and is trained using multiple mentions simultaneously. We initialise the mention embedding Transformer model weights from the RoBERTa \cite{roberta} model and train our model for 1M steps with a batch size of 64 and a maximum sequence length of 512 tokens. This requires approximately 4 days when using 8 V100 GPUs. For additional details, see Appendix \ref{sec:training-details}.

\section{Results}

\begin{table*}[h]

\centering
\resizebox{.92\linewidth}{!}{%
\begin{tabular}{@{}lccccccc@{}}
\toprule
\textbf{Method}                                        & \textbf{AIDA}  & \textbf{MSNBC}      & \textbf{AQUAINT}    & \textbf{ACE2004}    & \textbf{CWEB} & \multicolumn{1}{c|}{\textbf{WIKI}}       & \textbf{AVG} \\ \midrule
AIDA {{\cite{aida}}} & 78.0           & 79.0                & 56.0                & 80.0                & 58.6          & \multicolumn{1}{c|}{63.0}                & 69.1          \\
TagMe 2 \cite{Ferragina2012FastAA}                 & 70.6           & 76.0                & 76.3                & 81.9                & 68.3          & \multicolumn{1}{c|}{-}                   & -             \\
REL \cite{rel}                           & 89.4           & 90.7                & 84.1                & 85.3                & 71.9          & \multicolumn{1}{c|}{73.1}                & 82.4          \\
GENRE \cite{decao2020autoregressive}                            & {93.3$^{**}$}    & {\ul 94.3}          & 89.9                & 90.1                & 77.3          & \multicolumn{1}{c|}{{\ul 87.4}}          & {\ul 88.7}    \\
Bootleg$^{*}$ \cite{bootleg}                                         & 80.9           & 80.5                & 74.2                & 83.6                & 70.2             & \multicolumn{1}{c|}{76.2}                   & 77.6             \\
WNEL \cite{le-titov-2019-boosting}                                     & 89.7           & 92.2                & {\ul 90.7}          & 88.1                & {78.2}    & \multicolumn{1}{c|}{81.7}                & 86.8          \\
RLEL \cite{fang}                                     & \textbf{94.3$^{**}$} & 92.8                & 87.5                
& {\ul 91.2}          & {\ul {78.5}} & \multicolumn{1}{c|}{82.8}                & 87.9          \\
DCA-RL  \cite{yang-etal-2019-learning}                                     & {\ul{93.7$^{**}$}} & 93.8                & 88.3                
& {90.1}          & 75.6 & \multicolumn{1}{c|}{78.8}                & 86.7          \\
BiBSG  \cite{yang-etal-2018-collective}                                     & {{93.0$^{**}$}} & 92.6                & 89.9                
& {88.5}          & \textbf{81.8}  & \multicolumn{1}{c|}{79.2}                & 87.5          \\

\midrule
\textbf{KBED}                                          & 90.4           & {\textbf{94.8}} & {\textbf{92.6}} & {\textbf{93.4}} & {78.2}    & \multicolumn{1}{c|}{{\textbf{90.4}}} & \textbf{90.0} \\ \midrule
\multicolumn{8}{c}{\textbf{Model Ablations}}                                                                                                                                                                          \\ \midrule
\textbf{w/o KB}            & 87.5              & 94.4                   & 91.8                   & 91.6                   & 77.8             & \multicolumn{1}{c|}{88.7}                   & 88.6             \\
\textbf{KB only}                                       & 80.3              & 88.9                   & 83.0                   & 85.0                   & 69.7             & \multicolumn{1}{c|}{80.8}                   & 81.3             \\
\textbf{Entity types only}                      & 85.7              & 91.8                   & 91.8                   & 89.8                   & 74.3             & \multicolumn{1}{c|}{86.1}                   & 86.6             \\

\textbf{Entity descriptions only}                             & 84.8              & 90.5                  & 91.8                   & 90.8                   & 74.1             & \multicolumn{1}{c|}{87.7}                   & 86.6             \\ 

\textbf{Bilinear RE layer} & 86.5 & 94.4 & 91.4 & 93.6 & 77.5 & \multicolumn{1}{c|}{90.9} &  89.1 \\

\bottomrule
\end{tabular}
}
\caption{Entity disambiguation InKB micro F1 scores on test sets. The best value (excl. model ablations) is \textbf{bold} and second best is {\ul underlined}. $^{*}$We produced results using the code released by the authors. 
$^{**}$Indicates the model was trained on both AIDA and Wikipedia hyperlinks.}
\label{tab:standard-ed}
\end{table*}


\subsection{Standard ED}
The results in Table \ref{tab:standard-ed} show our model (KBED) achieves the highest average performance across the datasets by a margin of 1.3 F1, reducing errors by 11.5\%. The ablation results indicate the majority of the improvements across the datasets are attributable to our novel KB module. We observe the largest improvement of 3.0 F1 on the WIKI dataset, which is likely due to the documents having high factual density, enabling our model to leverage more KB facts (see Section \ref{sec:analysis} for relation analysis). Despite our model only be trained on Wikipedia, we obtain competitive results on out-of-domain datasets, such as MSNBC news articles, which implies the patterns learned from Wikipedia are applicable to other domains. In addition, the results demonstrate that our 3 modules (entity typing, entity descriptions, and KB facts) are complementary; when any module is used in isolation it reduces performance, demonstrating the benefits of a multifaceted approach to ED. Surprisingly, when our KB module is used in isolation it performs on par with the TagMe baseline, which suggests there is reasonable overlap between KB facts and the facts predicted from documents. Note that the AIDA results in Table \ref{tab:standard-ed} contain a mixture of models fine-tuned on this dataset (denoted with **) and trained on Wikipedia only (as in our case), so the numbers are not directly comparable.


\subsection{Long-tail and ambiguous ED}
Our model achieves an average F1 score of 70.1 on the original ShadowLink dataset (Table \ref{tab:tail-ed}) which substantially outperforms (+16.5 F1) embeddings-based models (GENRE, REL) and moderately outperforms (+4.0 F1) the Bootleg model \cite{bootleg} which is optimised for tail-performance and also uses entity types and KB facts. On the original dataset, the impact of our KB module is negligible because the limited document context reduces the chances of KB-related entities co-occurring; the strong performance is therefore largely due to the combination of entity types and descriptions. However, we see a notable average improvement of 12.7 F1 on the document-level version of the dataset, with the KB module having a considerable impact especially on the overshadowed entity subset where it contributes 6.7 F1. The performance margin between our model and Bootleg is greater when document-level context is used likely because Bootleg is designed for short contexts and has limited control over which KB facts to use for disambiguation, as all facts are weighted uniformly. We include a more extensive model ablation study in Appendix \ref{sec:ablation-study}.

\begin{table*}[h]

\centering
\resizebox{.92\linewidth}{!}{%
\begin{tabular}{@{}lccccccc@{}}
\toprule
\textbf{Method}                          & \textbf{SHADOW} & \textbf{TOP}  & \multicolumn{1}{c|}{\textbf{TAIL}} & \multicolumn{1}{c|}{\textbf{AVG}}  & \textbf{SHADOW-DOC} & \multicolumn{1}{c|}{\textbf{TOP-DOC}} & \textbf{DOC-AVG} \\ \midrule
AIDA {{\cite{aida}}}                & 35              & 56            & \multicolumn{1}{c|}{67}            & \multicolumn{1}{c|}{52.7}          & -                   & \multicolumn{1}{c|}{-}                & -                \\
TagMe 2 \cite{Ferragina2012FastAA} & 29              & 57            & \multicolumn{1}{c|}{83}            & \multicolumn{1}{c|}{56.3}          & -                   & \multicolumn{1}{c|}{-}                & -                \\
GENRE$^{*}$ \cite{decao2020autoregressive}              & 26            & 42          & \multicolumn{1}{c|}{93}            & \multicolumn{1}{c|}{53.7}          & 40.9                   & \multicolumn{1}{c|}{59.2}                &      50.1           \\
REL \cite{rel}             & 21              & 54            & \multicolumn{1}{c|}{91}            & \multicolumn{1}{c|}{55.3}          & -                   & \multicolumn{1}{c|}{-}                & -                \\
Bootleg$^{*}$ \cite{bootleg}                                  & 44.5               & 60.0             & \multicolumn{1}{c|}{93.7}             & \multicolumn{1}{c|}{66.1}             & 46.9                   & \multicolumn{1}{c|}{62.7}                & 54.8                \\ \midrule
\textbf{KBED}                            & \textbf{47.6}   & \textbf{64.2} & \multicolumn{1}{c|}{\textbf{98.5}} & \multicolumn{1}{c|}{\textbf{70.1}} & \textbf{60.8}       & \multicolumn{1}{c|}{\textbf{74.2}}    & \textbf{67.5}    \\ \midrule
\multicolumn{8}{c}{\textbf{Model Ablations}}                                                                                                                                                                                          \\ \midrule
\textbf{w/o KB}                          & {46.4}      & {64.2} & \multicolumn{1}{c|}{{98.3}}    & \multicolumn{1}{c|}{{ 69.6}}    & 54.1                & \multicolumn{1}{c|}{72.0}               & 63.1             \\
\textbf{KB only}                          & {26.9}      & {45.7} & \multicolumn{1}{c|}{{98.4}}    & \multicolumn{1}{c|}{{ 57.0}}    & 41.9                & \multicolumn{1}{c|}{60.0}               & 51.0             \\

\textbf{Entity descriptions only}                          & {42.1}      & {54.7} & \multicolumn{1}{c|}{{97.8}}    & \multicolumn{1}{c|}{{ 64.9}}    & 52.6                & \multicolumn{1}{c|}{65.0}               & 58.8             \\

\textbf{Entity types only}                          & {39.6}      & {55.6} & \multicolumn{1}{c|}{{98.5}}    & \multicolumn{1}{c|}{{ 64.6}}    & 47.3                & \multicolumn{1}{c|}{62.1}               & 54.7             \\

\textbf{Bilinear RE layer}  & 47.1 & 61.5 & \multicolumn{1}{c|}{{97.7}}  & \multicolumn{1}{c|}{{ 68.8}} & 59.2 & \multicolumn{1}{c|}{72.7} & 66.0 \\

\bottomrule
\end{tabular}
}
\caption{Entity disambiguation InKB micro F1 scores on ShadowLink test sets. SHADOW-DOC and TOP-DOC refers to the extended version of the dataset which includes the full-text of the document to use as additional context. The best value is \textbf{bold}. $^{*}$We produced results using the code released by the authors.}
\label{tab:tail-ed}
\end{table*}

\subsection{Relation extraction module}

To analyse the impact of the doc-level RE architecture introduced in Section \ref{subsec:relation-extraction} we present results in Tables \ref{tab:standard-ed} and \ref{tab:tail-ed} of performance with a standard bilinear RE layer \cite{ssan}. Our RE architecture leads to an average increase of 0.9 F1 on the standard ED datasets, of 1.3 F1 on the standard ShadowLink splits, and of 1.5 F1 on the ShadowLink doc-level splits. In addition, by avoiding the quadratic complexity bilinear layer, we achieve an increase in inference speed of approximately 2x, as measured on AIDA documents. We include doc-level RE results for our architecture on the DOCRED \cite{docred} dataset in Appendix \ref{sec:doc_level_re}.

\subsection{Error Analysis}

In Table \ref{tab:errors} we show the results from annotating 50 examples in which the model made an incorrect prediction for both the AIDA test split and the ShadowLink {SHADOW-DOC} split. \textbf{Gold not in cands.} refers to cases in which the gold entity was not in the top-30 candidates from the PEM table; \textbf{Missing KB fact} are cases where the model correctly predicted a relation connecting two mentions, but the corresponding fact was not in the KB; \textbf{Dominant PEM} is when the initial PEM score for one candidate was high (> 0.8), and the model fails to override this score; \textbf{Incorrect RE pred.} are cases in which the model makes an incorrect RE prediction between two mentions, and where this wrong prediction leads to the wrong choice of entity; \textbf{Ambiguous ann.} refers to gold annotations that are either incorrect or ambiguous.\footnote{Note that some examples may contain more than one source of error (or contain an error not clearly in any category), so the sum of the rows will not necessarily be 50.}

\begin{table}[h]
\resizebox{.95\linewidth}{!}{%

\begin{tabular}{lcc}
\toprule
  & \textbf{AIDA} & \textbf{SHADOW-DOC} \\
  \midrule
\textbf{Gold not in cands.}  & 18 & 32 \\
\textbf{Missing KB fact} & 2 & 6 \\
\textbf{Dominant PEM} & 0 & 1 \\
\textbf{Incorrect RE pred.} & 1 & 2 \\
\textbf{Ambiguous ann.} & 24 & 4 \\
\midrule
\textbf{Total} & 50 & 50 \\
\bottomrule
\end{tabular}
}
\caption{Counts per error category from 50 annotations on AIDA-CoNLL and ShadowLink-Shadow datasets.}
\label{tab:errors}
\end{table}

The results in Table \ref{tab:errors} indicate that the largest source of error is the gold entity not being present in the top-30 candidates. This is particularly true for the ShadowLink {SHADOW-DOC} split, as this split contains a larger number of tail entities which are less likely to be mentioned on Wikipedia. For the AIDA dataset, there are also many cases which are in some sense ambiguous.\footnote{These are often cases with national sports teams, such as ``Little will miss Australia's fixture...'' where ``Australia'' could refer to the country Australia or the Australian rugby team.} There are 8 cases in total where the model predicts a relation which it expects to be in the KB, but which is not in fact present. This is largely in the ShadowLink split, where tail entities are likely to be less well represented in Wikidata. The model is generally good at not depending on entity priors; despite every gold candidate in the Shadowlink SHADOW-DOC split being ``overshadowed'' by a more popular entity in the PEM table, there is only one example where the model fails to override this. Although the model often ``over-predicts'' relations between mentions, it rarely gets penalised for doing so, as in general the extra facts it predicts are not in the KB, meaning the \textbf{Incorrect RE pred.} count is low.

To further explore the role of missing candidates, Table \ref{tab:gold-percentage} shows the percentage of the gold entities present in the top-30 candidates we pass to the model, representing a hard upper-bound on the recall our model can achieve. The results vary from a high coverage of 99.5 for the MSNBC dataset, which largely contains head entities, to a lower coverage for the ShadowLink SHADOW (75.3) and TOP (83.6) splits. Table \ref{tab:gold-percentage} also shows the coverage if we pass all PEM candidates to the model. For some datasets, such as WIKI, this increases the coverage significantly. However, for the ShadowLink SHADOW split, the coverage is still below 80\%, indicating that better candidate generation strategies are an interesting avenue for future research. 


\begin{table}[h]
\resizebox{.95\linewidth}{!}{%
\setlength\tabcolsep{3pt}
\begin{tabular}{ccccccc}
\toprule
\multicolumn{3}{l}{\textbf{Main datasets test splits}} & &  \\
\textbf{n} & \textbf{AIDA}  & \textbf{MSNBC}      & \textbf{AQUAINT}    & \textbf{ACE2004}    & \textbf{CWEB} & \textbf{WIKI} \\
\midrule
\textbf{30} & 97.8 & 99.5 & 95.1 & 90.9 & 95.9 & 93.7 \\
\textbf{All} & 1.0 & 99.5 & 95.8 & 92.9 & 97.0 & 98.1 \\
\toprule
\multicolumn{3}{l}{\textbf{ShadowLink splits}}  & & & \\
 & \multicolumn{2}{c}{\textbf{SHADOW}}  & \textbf{TOP}      & \textbf{TAIL} & & \\
\midrule
\textbf{30} & \multicolumn{2}{c}{75.3} & 83.6 & 98.6 & &\\
\textbf{All} & \multicolumn{2}{c}{76.8} & 84.3 & 98.7 & &\\

\bottomrule
\end{tabular}
}
\caption{Percentage of gold entities in top-n candidates by dataset. We set n=30 for this paper.}
\label{tab:gold-percentage}
\end{table}


\section{Analysis}

\subsection{Relation predictions}
\label{sec:analysis}
To understand the relations which the model utilises to make predictions, Table \ref{tab:relation-analysis-wiki} displays for the WIKI dataset the number of KB (Wikidata) facts which exist between gold annotated mentions in the documents (\textbf{Gold}), the number of facts between mentions our model predicts with a score above 0.5 (\textbf{Predicted}) and the percentage of gold facts which our model also predicts (\textbf{Recall}).\footnote{Note that as the RE predictions are continuous, the \textbf{quantity} of facts our model predicts depends entirely on the choice of this threshold.}

The \emph{SAME AS} relation is used extensively by the model, demonstrating that using coreferences to other (potentially easier to disambiguate) mentions of the same entity in the document is a powerful addition for ED. We leave evaluation of the model on the coreference-specific task to future work. The \emph{OTHER} relation is also commonly predicted, suggesting the long tail of relations in Wikidata still hold useful information. The other widely used relations are generally either geographical or sports related, which is expected given the large number of sports entities in Wikidata. 

The recall numbers appear low, although this is expected behaviour in that the existence of a \textbf{Gold} fact does not necessarily imply that the text in the document infers this fact. For example, the text ``Donald Trump visited New York'' would include the gold fact [Donald Trump] [place of birth] [New York] but making this prediction for all sentences of this form would likely harm performance.

\begin{table}[h]
\resizebox{.95\linewidth}{!}{%

\begin{tabular}{lccc}
\toprule
  & \textbf{Gold} & \textbf{Predicted} & \textbf{Recall} \\
  \midrule
sport                                            &  1083 &   1028 &  0.53 \\shares border with                               &  1012 &   5211 &  0.68 \\\emph{OTHER}                                              &  1011 &  10077 &  0.29 \\\emph{SAME AS}                                            &   940 &   9666 &  0.36 \\country                                          &   890 &    285 &  0.10 \\located in the a.t.e &   709 &   4912 &  0.66 \\contains a.t.e       &   278 &    319 &  0.26 \\instance of                                      &   151 &   1241 &  0.23 \\country of citizenship                           &   120 &     90 &  0.08 \\subclass of                                      &    90 &   2430 &  0.26 \\genre                                            &    83 &    204 &  0.29 \\part of                                          &    80 &   2154 &  0.44 \\follows                                          &    69 &   1104 &  0.67 \\followed by                                      &    68 &   1312 &  0.71 \\member of sports team                            &    63 &    449 &  0.83 \\
\bottomrule
\end{tabular}
}
\caption{Analysis of relation predictions for WIKI dataset with threshold 0.5. 320 documents with 6772 entity mentions.}
\label{tab:relation-analysis-wiki}
\end{table}

\section{Conclusion}
We presented a novel ED model, which achieves SOTA performance on well-established ED datasets by a margin of 1.3 F1 on average, and by 12.7 F1 on the challenging ShadowLink dataset. These results were achieved by introducing a method to incorporate large symbolic KB data into an ED model in a fully differentiable and scalable fashion. Our analysis shows that better candidate-generation strategies are an interesting avenue for future research, if results are to be pushed higher on ambiguous and tail entities. Dynamic expansion of the KB by incorporating facts identified by the ED model is also a potentially promising direction.


\section*{Acknowledgements}

We would like to thank Vera Provatorva for providing us with the extended version of the ShadowLink dataset and Laurel Orr for assisting us with running the Bootleg baseline.

\clearpage


\bibliography{anthology,custom}
\bibliographystyle{acl_natbib}

\clearpage

\appendix

\section{Entity Type Selection}
\label{sec:model-details}
Our entity types are formed from direct Wikidata relation-object pairs and relation-object pairs inferred from the Wikidata subclass hierarchy; for example, (instance of, geographical area) can be inferred from (instance of, city). We only consider types with the following relations: instance of, occupation, country, and sport. We select types by iteratively adding types that separate (assuming an oracle type classifier) the gold entity from negative candidates for the most examples in our Wikipedia training dataset.

\section{Training Details}
\label{sec:training-details}
We use the Hugging Face implementation of RoBERTa \cite{huggingface} and optimise our model using Adam \cite{adam} with a linear learning rate schedule. We ignore the loss from mentions where the gold entity is not in the candidate set. Our model has approximately 197M trainable parameters. We present our main hyperparameters in Table \ref{tab:hyperparams}. Due to the high computational cost of training the model, we did not conduct an extensive hyperparameter search. To reduce GPU memory usage to below 16 GB during training, we subsample 30 mentions per context window, and subsample 5 candidates per mention (subsampling is not required during inference).

\begin{table}[h]
\centering
\resizebox{.8\linewidth}{!}{
\begin{tabular}{@{}ll@{}}
\toprule
\textbf{Hyperparameter} & \textbf{Value} \\ \midrule
learning rate           & 3e-5           \\
batch size              & 64             \\
max sequence length     & 512            \\
dropout                 & 0.05           \\
task hidden layer units      & 768            \\ 
\# training steps                & 1M             \\
\# candidates           & 30             \\
\# relations            & 128            \\
\# entity types      & 1400            \\ 

mention transformer init.  & roberta-base \\
\# mention encoder layers     & 12            \\
description transformer init.  & roberta-base \\
\# description encoder layers     & 2            \\ 
\# description tokens     & 32            \\ 

RE transformer init.  & random \\
RE coarse-to-fine K     & 600            \\ 
\# RE transformer layers     & 4            \\ 
\bottomrule
\end{tabular}
}
\caption{Our model hyperparameters.}
\label{tab:hyperparams}
\end{table}

\section{Model Ablation Study}
\label{sec:ablation-study}
In this section, we measure the contribution of key aspects of our model. For each model ablation, we train our model from scratch on the AIDA-CoNLL training set and evaluate on the development set, keeping hyperparameters constant. Surprisingly, the performance of our model is strong in this limited data setting, which means that our model is not dependent on a large set of training examples when there is a small amount of annotated in-domain data. Note that for ``w/o 128 standard relations'' we collapse all standard relations into the \emph{OTHER} special relation; and for ``w/o RE transformer'' we replaced the RE transformer with a single bilinear layer. 
\begin{table}[h]
\centering
\resizebox{.8\linewidth}{!}{
\begin{tabular}{@{}lc@{}}
\toprule
\textbf{Method}                               & \textbf{AIDA} \\ \midrule
\textbf{KBED}                                 &              \textbf{94.37} \\ \midrule
{w/o KB}               &               \textcolor{red}{92.20} \\
{w/o \emph{SAME AS} relation}               &               \textcolor{red}{93.65} \\
{w/o \emph{OTHER} relation}                 &               94.23 \\
{w/o 128 standard relations (collapsed)}           &               \textcolor{red}{93.67} \\
{w/o RE transformer}                   &               94.06 \\
{w/o weighting facts by entity scores} &               94.23 \\
{w/o weighting facts by relation scores} &               93.44 \\
{w/o reflexive RE} &               93.84 \\
{w/o entity descriptions}              &               93.89\\
{w/o entity types}                     &               \textcolor{red}{93.47} \\
{w/o entity priors}                    &               \textcolor{red}{93.63} \\
{w/o task hidden layers}                    &               94.07 \\
{w/o negative relation scores}                    &               94.19 \\
{with Wikipedia ED pre-training}                    &               95.58 \\ \bottomrule
\end{tabular}
}
\caption{ED F1 score on AIDA-CoNLL development split for model ablations trained from scratch on AIDA-CoNLL training split using the standard CoNLL candidates \cite{hoffart-etal-2011-robust}. The result is in \textcolor{red}{red} when the performance drops by more than 0.7.}
\label{tab:model-ablations}
\end{table}
Our results (Table \ref{tab:model-ablations}) indicate that all aspects of our model that we measured have a positive impact on performance. Interestingly, the KB module (+2.2 F1) has a greater impact than the entity description (+0.48 F1) and entity typing (+0.9 F1) modules despite weaker performance when used on its own (Table \ref{tab:standard-ed}). This implies there is less overlap between examples where KB module performs well, and the other modules perform well. We observe, the \emph{SAME AS} relation improves performance by 0.72 F1, which demonstrates that using coreference improves ED. Finally, we find that when the KB module has greater control over how to weight KB facts (based on the context) it leads to better results, for example if we collapse all standard relations into a single relation our performance drops by 0.7 F1.

\section{Doc-level RE results on DOCRED}
\label{sec:doc_level_re}

To verify the performance of our document-level RE architecture introduced in Section \ref{subsec:relation-extraction}, we present results of models trained and evaluated on the DOCRED dataset \cite{docred}. Our baseline implementation uses roberta-base as an encoder and a bilinear output layer. We show two variants in Table \ref{tab:docred}, a bilinear layer with input dimension 128 and with input dimension 256, which give an F1 score of 57.8 and 58.4 respectively. This compares to a score of 59.5 for an equivalent baseline implemented in \cite{ssan}. The difference is explained by our baseline not giving the model access to the gold coreference information, which is allowed in the DOCRED task but which we exclude as it will not be available for our entity linking task. 

\begin{table}[h]
\centering
\resizebox{.9\linewidth}{!}{%
\begin{tabular}{lcc}
\toprule
 & F1 & \makecell{Train time \\ (seconds per epoch)}\\
\midrule
Baseline (bilinear)- SSAN imp. & 59.5 & - \\
SSAN (roberta-base) & 60.9 & - \\
\textbf{Baseline (bilinear-128)} & 57.8 & 155.7 \\
\textbf{Baseline (bilinear-256)} & 58.4 & 343.2 \\
\textbf{Coarse to fine (2 layers)} & 60.4 & 100.6 \\
\textbf{Coarse to fine (4 layers)} & \textbf{61.2} & 106.2 \\
\bottomrule
\end{tabular}
}
\caption{Document-level relation extraction F1 scores on the DOCRED dev dataset.}
\label{tab:docred}
\end{table}

Our coarse-to-fine approach, with 4 ``fine'' transformer layers, pushes the dev-level F1 up by 2.8 F1 to 61.2. This puts it slightly above the roberta-base version of the current state-of-the-art model, SSAN \cite{ssan}, which scores 60.9, and additionally has access to the gold coreference labels in the embedding layer of the model. This validates that our document-level RE architecture is capable of producing accurate relation predictions, which we see in the main results table (Table \ref{tab:standard-ed}) also translates into stronger ED performance. 

By avoiding the bilinear layer, our implementation is also faster to train, achieving 106.2 seconds per epoch on the DOCRED dataset on a single Tesla V100 GPU, compared to 155.7 seconds for the baseline model with a 128-dimension bilinear layer, and 343.2 seconds for the more accurate baseline model with a 256 dimension bilinear layer.

\section{Dataset details}
\label{sec:dataset_details}

\subsection{Dataset statistics}

We present the topic, number of documents and number of mentions for each dataset used for evaluation (Table \ref{dataset_statistics_ed}). The datasets used cover a variety of sources including wikipedia text, news articles, web text and tweets. Note that the performance of the model outside these domains may be significantly different. 

\begin{table}[h]
	\centering
 	\resizebox{.95\linewidth}{!}{
		\begin{tabular}{lccc}
		& \textbf{Topic} & \textbf{Num docs} & \textbf{Num Mentions} \\
		\toprule
		\textbf{AIDA} & news & 231 & 4464 \\
		\textbf{MSNBC} & news & 20 & 656\\
		\textbf{AQUAINT} & news & 50 & 743 \\
		\textbf{ACE2004} & news & 57 & 259\\
		\textbf{CWEB} & web & 320 & 11154\\ 
		\textbf{WIKI} & Wikipedia & 320 & 6821\\
		\textbf{ShadowLink-ALL} & web & 2712 & 2712\\
        \bottomrule
		\end{tabular}
		}
		\caption{Dataset statistics for entity disambiguation datasets.}
     	\label{dataset_statistics_ed}
\end{table}

\subsection{ShadowLink Full Text versions}

The authors of \citet{provatorova-etal-2021-robustness} kindly provided us with the full documents from which the shorter text snippets (usually one or two sentences) in the ShadowLink dataset were sourced. We were able to match 596 of the 904 examples in the SHADOW split to its corresponding document, and 530 out of the 904 examples in the TOP split. As some full articles were extremely long we limited the document-length to 10000 characters, centred around the single annotated entity. To validate that the subset of examples we were able to match to full documents were representative of the original dataset splits, we ran our model on the sentence-level versions of these subsets, achieving 47.7 on the SHADOW split (comparable to 47.6 in Table \ref{tab:tail-ed}) and 63.9 on the TOP split (comparable to 64.2 in Table \ref{tab:tail-ed}).

\section{Additional relation analysis}
\label{sec:additional-relation-analysis}

To expand on the analysis in Section \ref{sec:analysis} we also include the number of gold and predicted relations in documents in the AIDA dataset (Table \ref{tab:relation-analysis-aida}). The first clear difference is that there is a far higher count of gold \emph{SAME AS} facts in the AIDA dataset, which is potentially explained by pages on Wikipedia generally having hyperlinks for the first mention of an entity only.

It is also interesting to note that there are lower recall numbers for the AIDA dataset relative to WIKI (Table \ref{tab:relation-analysis-wiki}), indicating that the RE module may have ``overfit'' in some sense to the Wikipedia style of article, and may be less effective on AIDA style news articles. 

\begin{table}[h]
\centering
\resizebox{.9\linewidth}{!}{%

\begin{tabular}{lccc}
\toprule
  & \textbf{Gold} & \textbf{Predicted} & \textbf{Recall} \\
  \midrule
 \emph{SAME AS}                                              &  4410 &  14765 &  0.31 \\\emph{OTHER}                                             &  3169 &   5501 &  0.05 \\country                                          &  3066 &    728 &  0.06 \\member of sports team                            &   880 &   2653 &  0.44 \\country of citizenship                           &   861 &    150 &  0.02 \\shares border with                               &   628 &    294 &  0.01 \\member of                                        &   385 &    669 &  0.08 \\league                                           &   380 &    416 &  0.23 \\located in the a.t.e &   297 &   1199 &  0.09 \\contains a.t.e       &   261 &     11 &  0.01 \\headquarters location                            &   252 &   1924 &  0.15 \\country for sport                                &   175 &    130 &  0.00 \\has part                                         &   126 &    532 &  0.07 \\place of birth                                   &   123 &    365 &  0.11 \\sport                                            &   103 &     32 &  0.00 \\

\bottomrule
\end{tabular}
}
\caption{Analysis of relation predictions for AIDA dataset with threshold 0.5.}
\label{tab:relation-analysis-aida}
\end{table}

\section{Inference Speed and Scalability}
We measure the time taken to run inference on the AIDA-CoNLL test dataset and compare it to SOTA baselines. Table \ref{tab:inference-speed-test} shows the results alongside the average
ED performance on the 6 standard ED datasets (used in Table \ref{tab:standard-ed}). Our model is an order of magnitude faster than the baselines with comparable ED performance.

\begin{table}[h]
	\centering
	\resizebox{.9\linewidth}{!}{%
	\begin{tabular}{@{}lcc@{}}
		\toprule
		\textbf{Method}     & \textbf{Time taken (s)} & \textbf{Avg. ED F1}\\ \midrule
		\citet{Cao2020AutoregressiveER} & 2100 & 88.7 \\
		\citet{wu-etal-2020-scalable} bi-encoder    &   93  & 80.4 \\
		\citet{wu-etal-2020-scalable} cross-encoder & 917   & 87.2                  \\ 
		\citet{bootleg} & 438 & 77.6 \\ 
		\midrule
		\textbf{KBED}    & 96 & \textbf{90.0}           \\
		\textbf{w/o KB}    & \textbf{15} & 88.6           \\\bottomrule
	\end{tabular}
	}
	\caption{Time taken in seconds for EL inference on AIDA-CoNLL test dataset.}\label{time_taken}
	\label{tab:inference-speed-test}

\end{table}

The most computationally expensive part of our model (accounting for approximately 80\% of the inference and training time) is computing the KB score due to the large number of pairwise interactions present in documents. The hyperparameter for coarse-to-fine relation extraction can be lowered to trade-off computation cost with ED performance by reducing the number of pairwise interactions. Alternatively, as computation of the initial entity score $\mathbf{\psi_a}$ is computationally cheap relative to the KB score $\mathbf{\psi_b}$, candidate entities with low initial entity scores can be pruned to further increase training and/or inference speed. These approaches would also allow scaling of the initial number of candidate entities to more than the 30 used for inference in this paper, if the use case required it.





\end{document}